\documentclass[sigconf]{acmart}
\usepackage{CJKutf8}

\usepackage[normalem]{ulem}
\useunder{\uline}{\ul}{}
\AtBeginDocument{%
  }

\acmConference[arXiv Preprint]{arXiv Preprint}

\begin{document}

\title{Towards Better Monolingual Japanese Retrievers with Multi-Vector Models}

\author{Benjamin Clavié}
\email{bc@answer.ai}
\affiliation{%
  \institution{Answer.AI}
  \city{Tokyo}
  \country{Japan}
}

\renewcommand{\shortauthors}{Clavié}

\begin{abstract}
As language-specific training data tends to be sparsely available compared to English, document retrieval in many languages has been largely relying on multilingual models. In Japanese, the best performing deep-learning based retrieval approaches rely on multilingual dense embedders, with Japanese-only models lagging far behind. However, multilingual models require considerably more compute and data to train and have higher computational and memory requirements while often missing out on culturally-relevant information. In this paper, we introduce JaColBERT, a family of multi-vector retrievers trained on two magnitudes fewer data than their multilingual counterparts while reaching competitive performance. Our strongest model largely outperform all existing monolingual Japanese retrievers on all dataset, as well as the strongest existing multilingual models on all out-of-domain tasks, highlighting the need for specialised models able to handle linguistic specificities. These results are achieved using a model with only 110 million parameters, considerably smaller than all multilingual models, and using only a limited Japanese-language. We believe our results show great promise to support Japanese retrieval-enhanced application pipelines in a wide variety of domains.
\end{abstract}



\maketitle
\section{Introduction}

Document retrieval has become a very important component of many applications, especially in the context of the rapid development of Retrieval-Augmented Generation (\textit{RAG}) applications, leveraging document retrieval to improve the capabilities of Large Language Models~\cite{rag} (\textit{LLMs}). Since the advent of the transformers architecture, deep-learning based approaches to document retrieval have become widely used~\cite{dpr}.

However, the focus has largely been on a handful of high-resources languages, such as English~\cite{mteb} and Mandarin~\cite{cpack}, with other languages receiving comparatively little attention. Large-scale corpora are largely lacking in most languages~\cite{mrtydi}, if they exist at all. As a result, the best-performing models on most lower-resources languages are multilingual models~\cite{m3}, which come with considerable downsides. Firstly, they are inefficient at the language-level, as highlighted by multilingual models requiring parameters count that are 3-to-5 times larger than their monolingual counterparts models to reach similar performance in high-resources languages such as English~\cite{me5, m3}. Moreover, they also showcase noticeable proficiency gaps between higher and lower resource languages~\cite{sparseresources} and recent work on LLMs has shown that even within high-resources languages, cultural specificities are often lost by multilingual models\cite{singaporecultural}.

For the Japanese language, there has historically been a lack of high-quality, large-scale training datasets~\cite{nagoya, jglue}, albeit there has recently been a growing local~\cite{jcse} and international efforts~\cite{miracl} to alleviate this issue, partially by relying on machine-translated English data~\cite{mmarco}. However, the use of machine translation can result in varying performances depending on the target language, which is especially apparent in languages that differ from English as much as Japanese~\cite{mt, mmarco}. As a result, mono-lingual Japanese retrievers, while more computationally efficient, have strongly lagged in performance in comparison to multilingual approaches\cite{m3}.

Recently, multi-vector retrieval methods such as ColBERT~\cite{colbert} have demonstrated impressive out-of-domain generalisation performance on many English-language retrieval tasks, while training on orders of magnitude fewer documents than similarly-performing single-vector models~\cite{colbertv2}. However, with the exception of certain cross-lingual pairs in relatively high resources language~\cite{colbertx}, there has been little efforts to develop multi-vector models for non-English languages. Finally, a further iteration on ColBERT, named ColBERTv2~\cite{colbertv2}, has shown that distilling knowledge from cross-encoders into a ColBERT model can further improve its generalisation potential.

\textbf{Contributions} In this work, we explore the intuition that multi-vector retrieval methods can serve to build a strong, low-parameter count mono-lingual retriever, in our case in Japanese, without the need for English-scale training corpora. To do so, we introduce the JaColBERT family of models, with JaColBERTv1 following the ColBERTv1~\cite{colbert} training approach and JaColBERTv2 introducing the ColBERTv2 knowledge distillation component~\cite{colbertv2}. Our models, respectively trained on two and one order of magnitude fewer data than state-of-the-art multilingual methods, vastly outperform all existing Japanese monolingual retrievers, as well as all existing models on out-of-domain benchmarks, and remains competitive on a task which is in-domain for other models.

Noticeably, our models achieve results that are competitive with multilingual models with 5 times the parameter count, showcasing much greater computational efficiency. We believe our results showcase the strong potential of multi-vector approaches for lower-resources languages retrieval, in particular for Japanese, and suggest potential improvements to be explore future research.

\section{Japanese Training Data}
\label{sec:data}

While there is a growing number of Japanese retrieval datasets \cite{jglue,jcse,nagoya, jacwir, jqara} there appears to have been a lack of high-quality large scale datasets to train generalist retrieval models. Recently, researchers have attempted to mitigate this issue by introducing MMARCO\cite{mmarco}, a machine-translated version of MS MARCO~\cite{msmarco}\footnote{MS MARCO is a widely used dataset for training retrieval models, often credited with spurring large performance gains on out-of-domain English benchmarks \cite{beir,m3,e5}.} into 13 languages. 

As a result, we choose to follow ColBERT\cite{colbert} and ColBERTv2\cite{colbertv2}, both trained on MS MARCO, and train the JaColBERT family on the Japanese split of MMARCO.

Recently, higher quality Japanese retrieval datasets have appeared with the release of Mr.TyDi \cite{mrtydi}, and an improved version of it with higher-quality annotations, MIRACL \cite{miracl}, two multilingual information retrieval datasets. However, Mr.TyDi\&MIRACL is also commonly used as an evaluation benchmark, and training on it greatly lowers the ability to evaluate the out-of-domain generalisation of multilingual models. As a result of this and compute limitations, we leave the exploration of Mr.TyDi and MIRACL as training corpora for future work.


Hard negatives are widely considered as a very important aspect of training retrieval models\cite{hardnegstudy}.  A \textit{hard negative} is a negative example that looks very similar to a positive example, and serves to improve a model's ability to discriminate between actually relevant passges and irrelevant passages that could appear relevant at first glance\cite{hardneg}. For both models, we employ hard negatives during training. We provide further details on the hard negative strategies used for training in Appendix~\ref{app:hardnegs}.

\section{Method}

\subsection{JaColBERTv1}

JaColBERTv1 largely follows the ColBERTv1 training approach~\cite{colbert}.  Our model is trained using the official ColBERT codebase\footnote{\url{https://github.com/stanford-futuredata/ColBERT}}, using 10 million triplets sampled from MMARCO, as described in Section~\ref{sec:data} and following the hard negative strategy presented in Appendix~\ref{app:hardnegs}.

We initialise JaColBERTv1 from \textsc{bert-base-japanese-v3}\footnote{https://huggingface.co/tohoku-nlp/bert-base-japanese-v3}, a 110 million parameters Japanese variant of BERT. We experimented with smaller training runs using Japanese-RoBERTa and LUKE~\cite{luke}, and the results were consistently worse than initialising the model off BERT. We provide further information about the training setting for JaColBERTv1 in Appendix~\ref{app:trainingv1}.

Following ColBERT, both training and retrieval use the \textsc{MaxSim} scoring function~\cite{colbert}, and seek to maximise the score of (query, pos\_document) pairs in relation to (query, neg\_document) ones.

\subsection{Knowledge Distillation: JaColBERTv2}

We also introduce a second model, JaColBERTv2, which adds the ColBERTv2 knowledge distillation approach~\cite{colbertv2}.

Rather than training on \textit{(query, pos, neg)} triplets, we train this model using 32-way triplets. Effectively, for every individual query, we give the model a single positive example and 31 negative ones. This is coupled with \textbf{knowledge distillation}. Rather than training the model on absolute positive and negative documents, each labelled as simply \textsc{positive} or \textsc{negative}, we provide the model with scores for each $(query, document)$ pair, which the model attempts to learn using KL-Divergence Loss.

The intuition behind this is that such training allows the model to differentiate between high-scoring hard negatives which, due to the nature of automated hard negative mining, may be actually be false negatives, whose score will be high and particularly irrelevant documents, whose score will be low~\cite{improving}.

In order to perform knowledge distillation, we require relevance scores for each $(query, document)$ pair. Traditionally, these scores are obtained by running a cross-encoder across every pair, and using their score as the distillation objective~\cite{improving}. This is particularly costly, as the model must be ran on all pairs. Moreover, at the time of this work, there did not exist strong Japanese cross-encoder\footnote{As of this paper's submission time, the situation has considerably improved, with models such as \textsc{japanese-reranker-cross-encoder-large-v1} (\url{https://huggingface.co/hotchpotch/japanese-reranker-cross-encoder-large-v1}) reaching very strong performance. We leave this exploration to future work.}.

To alleviate this issue, we take advantage of the fact that MMarco\cite{mmarco}, 
is a direct translation of the English MS-Marco\cite{msmarco}. As a result, we leverage the distillation scores publicly released by the authors of ColBERTv2~\cite{colbertv2}. We directly use those scores, provided for the English corpus, by transposing them to their translated versions.

We provide further information in the training setting for JaColBERTv2 in Appendix~\ref{app:trainingv2}.

\section{Evaluation}
\label{sec:eval}

\begin{table*}[h]
\begin{center}
\begin{tabular}{lccccc|ccccc}
\multicolumn{1}{c}{}  & \multicolumn{5}{c|}{Multilingual Models}                                                                         & \multicolumn{5}{c}{Japanese-only Models}                                                                                        \\
\multicolumn{1}{c}{}  & \multicolumn{2}{c|}{bge-m3}                                   & \multicolumn{3}{c|}{multilingual-e5}             & \multicolumn{1}{c|}{GLuCoSE}        & \multicolumn{2}{c|}{sup-simcse-ja}             & \multicolumn{2}{c}{JaColBERT}            \\
\multicolumn{1}{c}{}  & dense          & \multicolumn{1}{c|}{all}                     & large          & base           & small          & \multicolumn{1}{c|}{}               & base              & \multicolumn{1}{c|}{large} & v1                & v2                   \\ \hline
JSQuAD                & 0.939          & \multicolumn{1}{c|}{0.958}                   & 0.953          & 0.934          & 0.934          & \multicolumn{1}{c|}{\textit{0.798}} & 0.793             & \multicolumn{1}{c|}{0.777} & 0.961             & {\ul \textbf{0.968}} \\
MIRACL                & \textit{0.728} & \multicolumn{1}{c|}{\textit{\textbf{0.752}}} & \textit{0.706} & \textit{0.647} & \textit{0.636} & \multicolumn{1}{c|}{0.348}          & 0.171             & \multicolumn{1}{c|}{0.199} & 0.583             & {\ul 0.667}          \\
JQaRA                 & 0.539          & \multicolumn{1}{c|}{0.576}                   & 0.554          & 0.471          & 0.492          & \multicolumn{1}{c|}{0.309}          & 0.312             & \multicolumn{1}{c|}{0.392} & 0.550             & {\ul \textbf{0.585}} \\
JaCWIR                & 0.864          & \multicolumn{1}{c|}{0.906}                   & 0.876          & 0.852          & 0.869          & \multicolumn{1}{c|}{0.686}          & 0.578             & \multicolumn{1}{c|}{0.474} & 0.904             & {\ul \textbf{0.919}} \\ \hline
Average               & 0.768          & \multicolumn{1}{c|}{\textbf{0.798}}          & 0.772          & 0.726          & 0.733          & \multicolumn{1}{c|}{0.535}          & 0.464             & \multicolumn{1}{c|}{0.461} & 0.75              & {\ul 0.785}          \\ \hline
Model size (relative) & \multicolumn{2}{c|}{5.11x}                                    & 5.03x          & 2.5x           & 1.06x          & \multicolumn{1}{c|}{1.20x}          & {\ul \textbf{1x}} & \multicolumn{1}{c|}{3.03x} & {\ul \textbf{1x}} & {\ul \textbf{1x}}   
\end{tabular}
\caption{Results on three retrieval tasks for each model. Results are reported using the metrics described in Section~\ref{sec:eval}. We also report the relative model size in terms of parameter count for each model, with JaColBERT serving as the reference point. Best overall results are reported in \textbf{bold}, and the best mono-lingual model results are \underline{underlined}. Results in \textit{italic} indicate the model was exposed to the task evaluated during its training and is therefore not evaluated out-of-domain.}
\label{results}
\end{center}
\end{table*}

\textbf{Data} We perform evaluation on a variety of datasets. We use the Japanese split \textbf{MIRACL}~\cite{miracl}, a general domain search dataset; JSQuAD~\cite{jglue}, a Japanese general-domain question answering; JQaRA\cite{jqara}, another QA dataset based on the JAQKET question-answering competition~\cite{JAQKET}; and JaCWIR~\cite{jacwir}, a Retrieval-Augmented Generation (RAG) evaluation dataset containing high-quality web pages and synthetic questions generated by OpenAI's GPT-3.5. Interestingly, while MIRACL is in-domain for all multilingual models, all other datasets are out-of-domain for all evaluated models, providing helpful metrics for potential real-world use cases.

\textbf{Metrics} Following standard practices, we use a wide array of metrics, depending on the dataset. We report NDCG@10 for MIRACL and JQaRA, Recall@3 for JSQuAD and MAP@10 for JaCWIR.

\textbf{Models} We compare our models with the best-existing Japanese-language retrievers, as well as the best multilingual models: the multilingual-E5 (mE5)~\cite{me5} family of embeddings and bge-m3~\cite{m3}. As bge-m3 is a multi-representations method, we report the results for both the \textit{dense} setting, using only its single-vector output, and \textit{all} setting, using a combination of dense, sparse and multi-vector representations.

It is worth noting that the majority of these models are considerably larger than JaColBERT models and trained on an order of magnitude more data. Further details on the retrieval settings are presented in Appendix~\ref{app:retrieval}.

\section{Results}

We present the evaluation results in Table~\ref{results}. Immediately, we notice that our models considerably outperforms all existing Japanese-only embedding approaches. While the previous best-performing Japanese model only reaches an average score of 0.535, with JaColBERTv1 reaching an average performance of \textbf{0.750}, and v2 reaching a score \textbf{0.785}.

These results show the considerable performance improvement obtained through knowledge distillation. These results are especially promising as, due to model availability and computational resources, we used the scores from a relatively weak cross-encoder. However, existing studies have shown a strong correlation between downstream results and the performance of the distilled cross-encoder~\cite{colbertx}, suggesting that scores from newer, stronger performing cross-encoders could improve our results further.

Interestingly, JaColBERT models are also particularly competitive with multilingual models. On all out-of-domain evaluation sets, they outperform all existing state-of-the-art models. On MIRACL, on which both bge-m3 and the  mE5 model family were trained, JaColBERTv2 outperforms the base and small variants of mE5, but yields weaker results than its large variant and bge-m3. 

However, it is worth noting that in order to reach these results, both of these models have been trained on an order of magnitude more data models and have more than 5 times the parameter count than  than JaColBERT models. This suggests that scaling our approach might allow us to at least partially close the gap without needing to train on in-domain data.

\section{Conclusion and Future Work}

In this work, we have introduced the JaColBERT family of multi-vector Japanese retrieval models. Building upon the recent advances in multi-vector retrieval, we show that the better performing of those models, JaColBERTv2, reaches state-of-the-art performance on a variety of retrieval datasets. It very strongly outperforms all pre-existing monolingual Japanese retrieval models on all evaluated tasks while also outperforming state-of-the-art multilingual models on all out-of-domain datasets.

Our results show that monolingual retrieval models can be highly competitive, despite comparatively limited resources in relation to English and being trained with an order of magnitude less training data and only a fifth of the parameters of multilingual retrievers. We believe that our work also supports the idea that multi-vector retrieval approaches are particularly well-suited to improving performance in lower resources settings due to their strong generalisation potential, warranting further work in the area.

Moreover, while our models achieve strong results, our study has left many areas for future work, such as using better models to distill from, or leveraging LLMs for synthetic data generation.
We hope that our findings help support further research into stronger mono-lingual, and specifically Japanese-language, retrieval.

\begin{CJK*}{UTF8}{goth}
\bibliographystyle{ACM-Reference-Format}
\bibliography{bibliography}
\end{CJK*}


\appendix



\section{Hard Negatives}
\label{app:hardnegs}

\subsection{JaColBERTv1}.
To support stronger Japanese retrieval models, we generate hard negatives for the MMARCO dataset, using two approaches:\\
\textbf{Dense Embeddings} We use the multilingual-e5-small model to retrieved the 110 most relevant documents for each query. We then discard the 10 highest scoring documents, as MMARCO is a lossy dataset: some passages for a query are not annotated as positive examples, although they could be considered relevant by a human reviewer. Finally, we randomly sample 25 example from the resulting 100 documents.

\textbf{BM25} We use the Anserini~\cite{anserini} implementation of BM25. For each individual query, we discard the ten highest-scoring documents before randomly sampling 10 documents from the remaining highest-scoring 100 documents.

The generated data is used to train JaColBERTv1, along with the initial training negatives provided in the original dataset. We make our full dataset available to support future work \footnote{\href{https://huggingface.co/datasets/bclavie/mmarco-japanese-hard-negatives}{https://huggingface.co/datasets/bclavie/mmarco-japanese-hard-negatives}}.

\subsection{JaColBERTv2}

We trained JaColBERTv2 using the MMARCO translation of a subsample of the MS-Marco triplets provided by the authors of ColBERTv2~\cite{colbertv2}. ColBERTv2 is trained using 19 million 64-way triplets. All of the negatives for each triplet were obtained via hard negative mining using BM25 and weak neural retrievers.

JaColBERTv2 uses a downsample of this data, using 8 million 32-way triplets. During our downsampling process, we attempted to keep all negative examples also present in the original JaColBERT hard negative set, and filled the rest of the 31 negatives with random sampling.

\section{Training Information}
\label{app:training}

\subsection{JaColBERTv1}
\label{app:trainingv1}

We train JaColBERT with a total batch size of 128 (16 per GPU). We perform training for a single epoch, iterating over the entire dataset once. The model is trained with 8000 warm-up steps, and a learning rate of 5e-6. Our experiments showed worse performance with other common learning rates, with 3e-6 being the closest. Learning rates of 1e-5 and 2e-5 resulted in noticeable performance degradation at early evaluation steps, and were not evaluated further. To use as training data, randomly sampled 10 million triplets from the hard negative training set described in Appendix~\ref{app:hardnegs}. Training was performed on 8 NVidia L40 GPU and took under 10 hours. 

\subsection{JaColBERTv2}
\label{app:trainingv2}

JaColBERTv2 is initialised off JaColBERTv1 and further trained on 8 million 32-way triplets. We trained with a total batch-size of 32 (4 per GPU) for 10000 warm-up steps with a learning rate of 1e-5, which reached stronger performance on development set evaluations in reduced-dataset ablation runs than alternative learning rates, for a single epoch. Training was conducted on 8 NVidia A100 40GB GPUs and took 28 hours.

\section{Retrieval Settings}
\label{app:retrieval}

For all baselines, we use exact search for all datasets except MIRACL, where we perform approximate-search using an HNSW index due to its size. The maximum document length is set to 512.

For JaColBERT models, We set the maximum query length to 64, and the maximum document length of 512. We perform exact search on 16-bit vectors for all datasets except MIRACL, for which we follow ColBERTv2 \cite{colbertv2} and apply 2-bit quantisation to all vectors.

\section{Fio Embeddings}

As an early precursor to this work, we also released \textsc{fio-base-japanese-v0.1} (Fio). Fio is initiated from \textsc{bert-base-japanese-v3},  trained for three epochs on JNLI \cite{jglue} and JSNLI \cite{nagoya}, then fine-tuned for a single epoch (due to compute constraints) on an extremely small subset (100 000 sentence pairs, half negatives and half positives) of MMARCO, as well as subsamples of MIRACL and Mr.TiDy. Fio is trained using AnglE optimisation \cite{angle}. More detail on Fio is outside the scope of this report and available on the associated \href{https://ben.clavie.eu/fio_v1}{release blog post}.

At this stage, Fio for retrieval remains a proof of concept and should not be used in lieu of JaColBERT or multilingual e5 models on these tasks. However, we believe that the data released with this work should allow to easily train a version a monolingual dense embedding model with strong retrieval performance, and intend to do so in the future.

\end{document}